\pdfoutput=1
\documentclass[11pt]{article}

\usepackage{ACL2023}

\usepackage{times}
\usepackage{latexsym}
\usepackage{graphicx}
\usepackage{color, colortbl}

\usepackage{float}
\restylefloat{table}

\usepackage{comment}
\usepackage{amsmath}
\usepackage[breakable]{tcolorbox}

\usepackage[T1]{fontenc}
\usepackage[utf8]{inputenc}

\usepackage{microtype}

\usepackage{inconsolata}

\definecolor{Gray}{gray}{0.9}

\newcommand*{\missingreference}{{\Huge \colorbox{red}{?reference?}}}
\newcommand*{\missingcitation}{{\Huge \colorbox{red}{?citation?}}}
\makeatletter
\def\@setref#1#2#3{%
  \ifx#1\relax
    \protect\G@refundefinedtrue
    \nfss@text{\reset@font\missingreference}%
    \@latex@warning{Reference `#3' on page \thepage \space
      undefined}%
  \else
    \expandafter#2#1\null
  \fi}
\def\@citex[#1]#2{\leavevmode
  \let\@citea\@empty
  \@cite{\@for\@citeb:=#2\do
    {\@citea\def\@citea{,\penalty\@m\ }%
      \edef\@citeb{\expandafter\@firstofone\@citeb\@empty}%
      \if@filesw\immediate\write\@auxout{\string\citation{\@citeb}}\fi
      \@ifundefined{b@\@citeb}{\hbox{\reset@font\missingcitation}%
        \G@refundefinedtrue
        \@latex@warning
        {Citation `\@citeb' on page \thepage \space undefined}}%
      {\@cite@ofmt{\csname b@\@citeb\endcsname}}}}{#1}}
\makeatother

\title{Do CoNLL-2003 Named Entity Taggers Still Work Well in 2023?}
  
 \author{Shuheng Liu \and Alan Ritter \\
  School of Interactive Computing \\
  Georgia Institute of Technology \\
 \texttt{sliu775@gatech.edu, alan.ritter@cc.gatech.edu} 
\\}

\begin{document}
\maketitle
%In this paper, we introduce CoNLL++, a new annotated test set created in a similar way to the original CoNLL-2003 set, but using data from 2020. 
%We use CoNLL++ to evaluate the generalization of over 20 different models to modern data. Our results show that different models have varying levels of generalization. 
\begin{abstract}
%Named Entity Recognition (NER) is an important task, and 
The CoNLL-2003 English named entity recognition (NER) dataset has been widely used to train and evaluate
NER models for almost 20 years. However, it is unclear how well models that are trained on this
20-year-old data and developed over a period of decades using the same test set will perform
when applied on modern data. In this paper, we evaluate the generalization of over 20 different
models trained on CoNLL-2003, and show that NER models have very different generalization. 
Surprisingly, we find no evidence of performance degradation in pre-trained Transformers,
such as RoBERTa and T5, even when fine-tuned using decades-old data. 
We investigate why some models generalize well to new data while others do not,
and attempt to disentangle the effects of temporal drift and overfitting due to test reuse.
Our analysis suggests that most deterioration is due to temporal mismatch between the
pre-training corpora and the downstream test sets. We found that four factors are important
for good generalization: model architecture, number of parameters, time period of the
pre-training corpus, in addition to the amount of fine-tuning data.
We suggest current evaluation methods have, in some sense, underestimated progress
on NER over the past 20 years, as NER models have not only improved on the original
CoNLL-2003 test set, but improved even more on modern data.
Our datasets can be found at \url{https://github.com/ShuhengL/acl2023_conllpp}.
% We will release the datasets created.
\end{abstract}

\section{Introduction}

\begin{figure}[ht]
    \centering
    \includegraphics[scale=0.6]{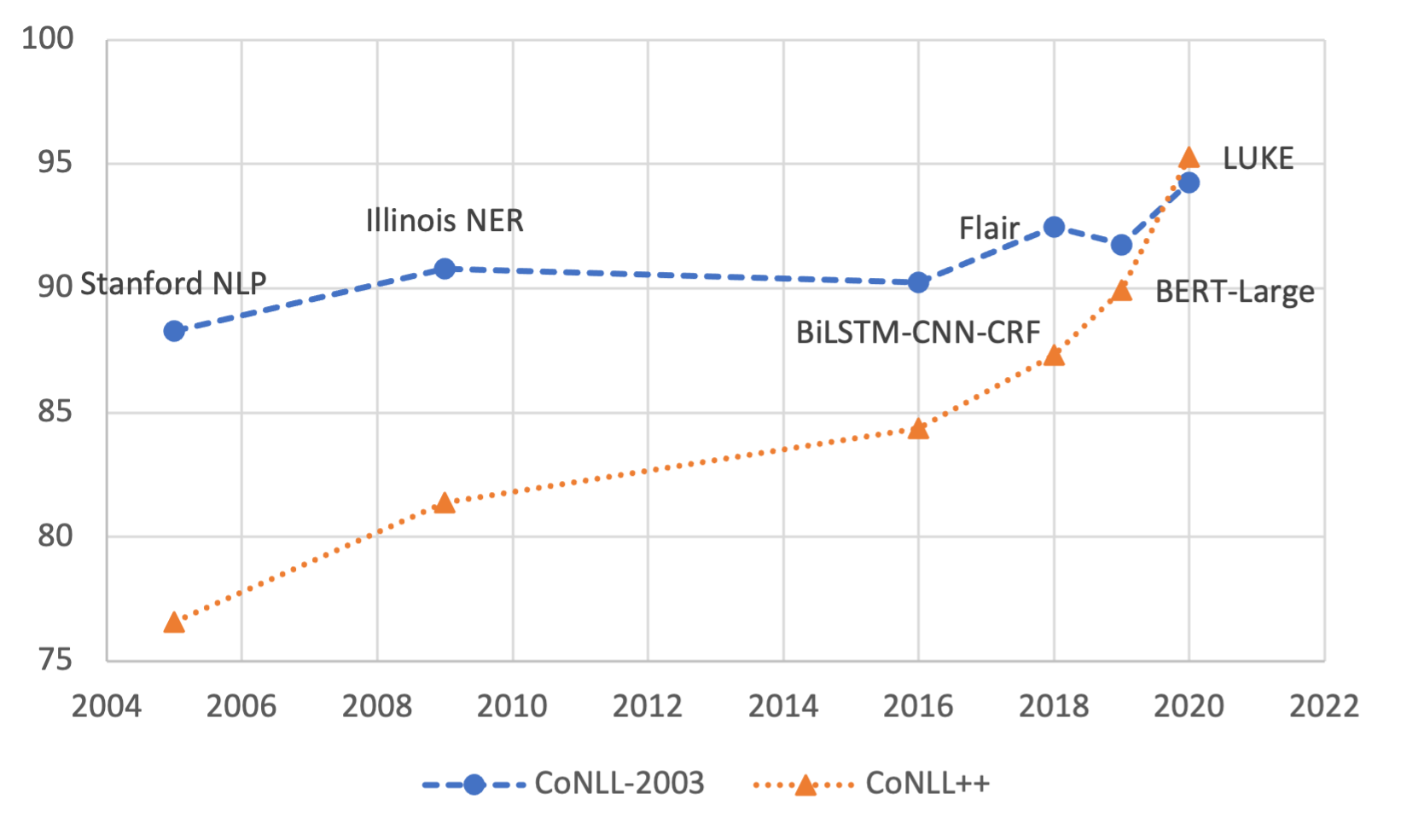}
    \caption{Progress on NER from 2005-2020, as measured using published F\textsubscript{1} scores on the CoNLL-2003 (data from 1996)
    and CoNLL++ (data from 2020) English NER test set. The gap between the two grows smaller as time
    passes by, showing improved generalization of models developed over time.}
    \label{fig:timeline}
\end{figure}

\begin{table*}[ht!]
  \centering
  \small
  \begin{tabular}{|l|lccccccc|}
  \hline \textbf{Dataset} & \textbf{Time} & \textbf{\# Tokens} & \textbf{LOC} & \textbf{MISC} & \textbf{ORG} & \textbf{PER} & \begin{tabular}{@{}c@{}} \textbf{\# Unique} \\ \textbf{Tokens} \end{tabular} & \begin{tabular}{@{}c@{}} \textbf{Avg Sentence} \\ \textbf{Length} \end{tabular} \\
  \hline
  CoNLL-2003 & Dec. 1996 & 46,435 & 1668 & 702 & 1661 & 1617 & 9,489 & 12.67\\
  \hline
  CoNLL++ & Dec. 2020 & 46,587 & 1128 & 697 & 1201 & 981 & 8,115 & 23.64 \\
  \hline
  \end{tabular}
  \caption{\label{tab:stat} Statistics of CoNLL-2003 test set and our CoNLL++.
  We report the publication time of the articles,
  the numbers of four different types of entities,
  the number of tokens, unique tokens and average number of tokens per sentence.\footnotemark[1]}
\end{table*}

The progress of natural language processing (NLP) is typically measured using performance metrics like accuracy or F\textsubscript{1} score on public test sets.
For instance, the top line in Figure~\ref{fig:timeline} shows the steady improvement of selected models on the CoNLL-2003 English named entity recognition (NER) test set
% \citep{Sang:2003}
\citep{tjong-kim-sang-de-meulder-2003-introduction}
over the course of 15 years (2005-2020) as measured by published F\textsubscript{1} scores.

However, these scores are all calculated using the same publicly available test set, which raises several questions.
One concern is how much of this progress is actually due to \textit{adaptive overfitting}, i.e. over-estimating performance by reusing the same test set,
as opposed to genuine improvement \citep{recht2019imagenet,roelofs2019meta,gorman-bedrick-2019-need}.
In addition, there is also the issue of \textit{temporal drift} as training data ages,
which can negatively impact performance on modern data \citep{rijhwani-preotiuc-pietro-2020-temporally,agarwal-nenkova-2022-temporal,luu-etal-2022-time}.

Performance degradation is a significant concern in applications that use NER, such as text deidentification \citep{morris-etal-2022-unsupervised},
relation extraction \citep{zhong-chen-2021-frustratingly},
linking entities to a knowledge base \citep{de-cao-etal-2022-multilingual}, etc.
However, continuously annotating, training, and evaluating new models on new data is not always possible. NER models that are trained on decades-old data and evaluated on heavily-used public development and test sets may struggle to perform well on modern data, which highlights the need to consider these factors when assessing performance.

To understand how well NER works when models have been developed over 20 years using the same dataset,
we created a new test set called CoNLL++. We closely modeled CoNLL++ after the CoNLL-2003 test set,
using news articles from 2020 instead of 1996, as in the original dataset.
We carefully controlled for other variables, making results on the two datasets as comparable as possible, with the exception of the time frame.
An example of an annotated sentence from CoNLL++ is shown below:

% AMBASSADOR 4366 4376 O
% TO 4377 4379 O
% THE 4380 4383 O
% UNITED 4384 4390 I-ORG
% NATIONS 4391 4398 I-ORG
% : 4399 4400 O
% LINDA 4401 4406 I-PER
% THOMAS-GREENFIELD 4407 4424 I-PER
\begin{center}
  \begin{tabular}{ll}
    AMBASSADOR & O \\
    TO & O \\
    THE & O \\
    UNITED & I-ORG \\
    NATIONS & I-ORG \\
    : & O \\
    LINDA & I-PER \\
    THOMAS-GREENFIELD & I-PER
  \end{tabular}
\end{center}

Using CoNLL++, we conduct an empirical study of more than 20 NER models that were trained on the original CoNLL-2003
training split. 
Our analysis shows that different models can have very different generalization when moving to modern data.
%A model that appears to perform well according to a public benchmark from 2003 may or may not be as good on data from 2020. 
Simply comparing the performance of models on the
CoNLL-2003 test set does not tell the whole story of progress on NER over the past 20 years.

\begin{figure*}[ht]
    \centering
    \includegraphics[width=0.95\textwidth]{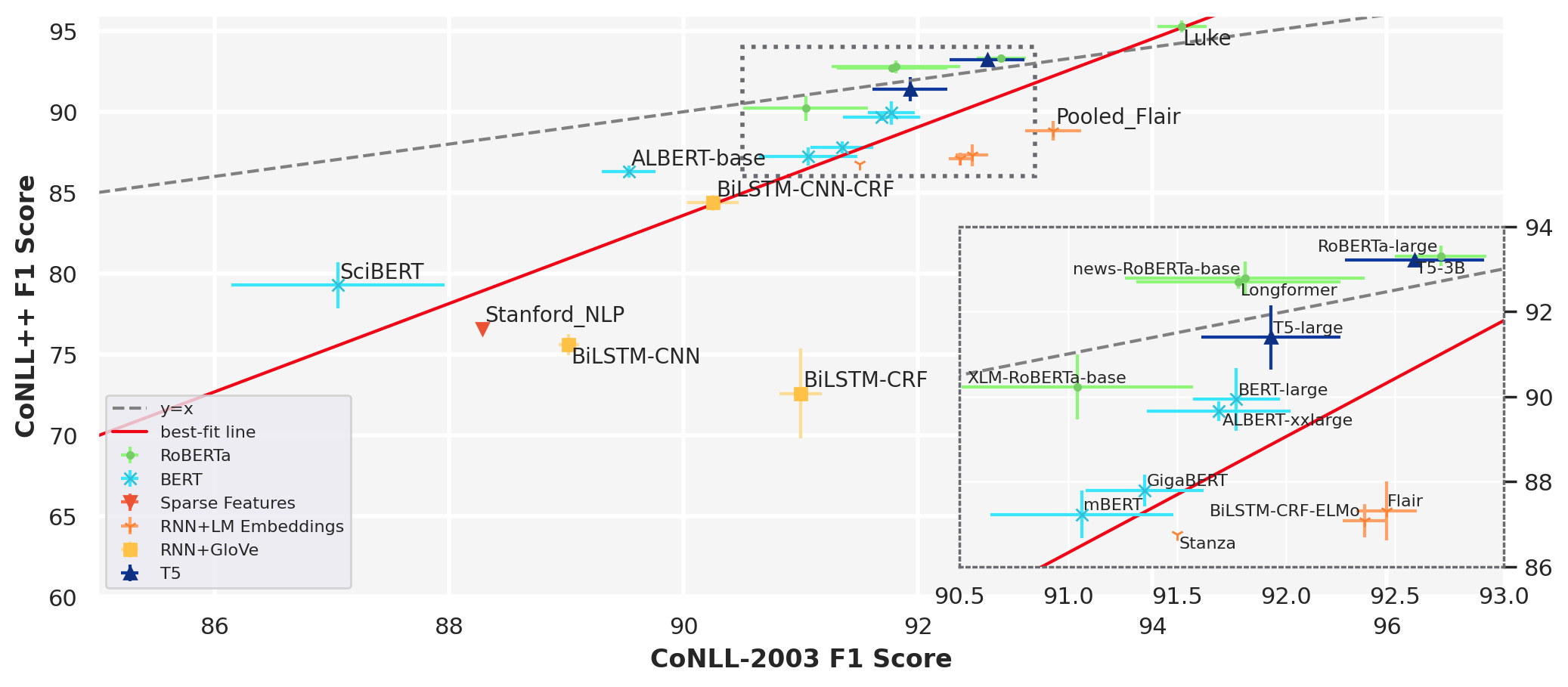}
    \caption{Plot of CoNLL++ F\textsubscript{1} scores against CoNLL-2003 F\textsubscript{1}
    scores. Each data point represents the average F\textsubscript{1} for each model, and the
    error bar represents one standard deviation. We observe that models show different level of generalization,
    while T5 and RoBERTa models generalize to CoNLL++. The solid best-fit line is steeper than the dashed $y=x$
    ideal generalization line, providing evidence against adaptive overfitting (\S~\ref{ssec:adaptive_overfitting}).
    This figure is best viewed in color.}
    \label{fig:results}
\end{figure*}

Similar to the findings of \citet{recht2019imagenet} on the ImageNet dataset \citep{deng2009imagenet},
we do not observe evidence of widespread overfitting on CoNLL-2003.  
%When considering a best-fit line of models' performance on the two datasets (Figure \ref{fig:results}), 
On average, each point of F\textsubscript{1} improvement on the CoNLL-2003 test set translates to a larger improvement on CoNLL++
(see Figure~\ref{fig:results}), suggesting overall improvements on the original dataset between 2003-2020 are mostly \textit{not} due to overfitting.
%When measuring diminishing returns on the new CoNLL++ test set, we do not find evidence for the hypothesis that models have overfit the CoNLL-03 test set after 17 years of heavy use, which is in-line with the findings of \citet{roelofs2019meta} on a public test set that has been in use for over 17 years.
Rather, most performance deterioration appears to be caused by temporal misalignment \citep{luu-etal-2022-time}.

Suprisingly, for some models (e.g. RoBERTa and T5), we find no evidence of performance degradation at all,
despite the fact they are fine-tuned
% and developed 
on a 20-year-old public dataset.
We conduct an extensive analysis, which suggests that
model size,
architecture,
amount of fine-tuning data,
and pre-training corpus
are all important factors for generalization in NER.

\footnotetext[1]{We notice that our dataset contains fewer entities than CoNLL-2003.
This is mainly because there are a number of tabular data, with information
such as results of sports events (e.g. 1. Jesper Ronnback ( Sweden ) 25.76 points),
in CoNLL-2003. Such data greatly contribute to the number of entities.
These tabular data also cause the average sentence length of CoNLL-2003
to be smaller than that of CoNLL++. By removing these data, we found that
the average sentence length increased to 18.50, much more
comparable to CoNLL++. Model perfomances reported in Figure~\ref{fig:results}
were also not affected by the removal of these tabular data. We include
further analysis and explanation in the Appendix (\S~\ref{sec:app_tabular_data}).
}

%\section{Data Collection}

\section{Annotating a New Test Set to Measure Generalization}

\noindent{\bf Data Collection:}
The CoNLL-2003 shared task collected English data from the Reuters Corpus, including Reuters news
articles published between Aug. 1996 and Aug. 1997. The test set was collected from December
1996 according to \citet{tjong-kim-sang-de-meulder-2003-introduction}. We find that almost all articles were published
between Dec. 5th and 7th, 1996, except one article published on Nov. 29th and another on Dec. 16th.
Our dataset follows this distribution to collect Reuters news articles published between December 5th
and 7th, 2020, collected from the Common Crawl Foundation.\footnote[2]{\url{http://commoncrawl.org/}}
We tokenize the data with the same tokenizer used for the CoNLL-2003 shared task, and randomly select
articles to match the total number of tokens in the original test set.

\begin{table*}[ht!]
  \resizebox{\textwidth}{!}{
      \centering
      \begin{tabular}{|l|l|l|c|r|c|c|}
          \hline
          \textbf{Name} & \textbf{Architecture} & \textbf{Reference} & \textbf{Corpus} & \textbf{Corpus Time} & $\mathbf{\Delta F_1}$ (\%) & $\mathbf{\Delta}$\textbf{Rank} \\
          \hline
          \rowcolor{Gray}
          BiLSTM-CRF & GloVe+RNN+CRF & \citet{lample-etal-2016-neural}\footnotemark[5] & WP & till 2014 & -20.25 & -7 \\
          BiLSTM-CNN & GloVe+RNN & \citet{chiu-nichols-2016-named}\footnotemark[6] & WP & till 2014 & -15.09 & 0 \\
          \rowcolor{Gray}
          Stanford NLP & CRF & \citet{finkel-etal-2005-incorporating} & - & - & -13.25 & 0 \\
          SciBERT & BERT & \citet{beltagy-etal-2019-scibert} & SS & till 2019* & -8.94 & +2 \\
          \rowcolor{Gray}
          BiLSTM-CNN-CRF & GloVe+RNN+CRF & \citet{ma-hovy-2016-end}\footnotemark[7] & WP & till 2014 & -6.52 & -1 \\
          % SpanBERT\textsubscript{Large} & BERT & \citet{joshi-etal-2020-spanbert} & BP & till 2018* & -6.52 & +2 \\
          BiLSTM-CRF-ELMo & ELMo+RNN+CRF & \citet{peters-etal-2018-deep} & 1B & till 2011* & -5.72 & -7 \\
          \rowcolor{Gray}
          Flair & GloVe+Flair+RNN+CRF & \citet{akbik-etal-2018-contextual} & WP & till 2014* & -5.57 & -7 \\
          \rowcolor{Gray}
          & & & 1B & till 2011* & & \\
          Stanza & Flair+RNN & \citet{qi-etal-2020-stanza} & 1B & till 2011* & -5.19 & -8 \\
          \rowcolor{Gray}
          Pooled Flair & GloVe+Flair+RNN+CRF & \citet{akbik-etal-2019-pooled} & WP & till 2014* & -4.65 & -6 \\
          \rowcolor{Gray}
          & & & 1B & till 2011* & & \\
          mBERT & BERT & \citet{devlin-etal-2019-bert} & WP & 01/2001-2018* & -4.22 & -1 \\
          \rowcolor{Gray}
          GigaBERT & BERT & \citet{lan-etal-2020-empirical} & G5 & 01/2009-12/2010 & -3.90 & 0 \\
          \rowcolor{Gray}
          & & & WP & till 2019* & & \\
          \rowcolor{Gray}
          & & & OS & 11/2018 & & \\
          ALBERT\textsubscript{Base} & BERT & \citet{Lan2020ALBERT:} & \textit{BP} & till 2018* & -3.61 & +3 \\
          \rowcolor{Gray}
          ALBERT\textsubscript{XXL} & BERT & \citet{Lan2020ALBERT:} & \textit{BP} & till 2018* & -2.22 & +4 \\
          BERT\textsubscript{Large} & BERT & \citet{devlin-etal-2019-bert} & BC & till 2015* & -2.01 & +3 \\
          & & & WP & 01/2001-2018* & & \\
          \rowcolor{Gray}
          XLM-RoBERTa\textsubscript{Base} & RoBERTa & \citet{conneau-etal-2020-unsupervised} & CC & till 2020* & -0.90 & +5 \\
          T5\textsubscript{Large} & Transformer & \citet{t5} & C4 & 04/2019 & -0.59 & +11 \\
          \rowcolor{Gray}
          RoBERTa\textsubscript{Large} & RoBERTa & \citet{roberta} & \textit{BP} & till 2018* & +0.64 & +3 \\
          \rowcolor{Gray}
          & & & CN & 09/2016-02/2019 & & \\
          \rowcolor{Gray}
          & & & OW & till 2019* & & \\
          \rowcolor{Gray}
          & & & ST & till 2018* & & \\
          T5\textsubscript{3B} & Transformer & \citet{t5} & C4 & 04/2019 & +0.67 & +1 \\
          \rowcolor{Gray}
          Longformer\textsubscript{Base} & RoBERTa & \citet{longformer} & \textit{BP} & till 2018* & +1.00 & +5 \\
          \rowcolor{Gray}
          & & & \textit{RP} & till 2019 & & \\
          \rowcolor{Gray}
          & & & ST & till 2018* & & \\
          \rowcolor{Gray}
          & & & RN & 12/2016-03/2019 & & \\
          news-RoBERTa\textsubscript{Base} & RoBERTa & \citet{gururangan-etal-2020-dont} & \textit{RP} & till 2019 & +1.06 & +5 \\
          & & & RN & 12/2016-03/2019 & & \\
          \rowcolor{Gray}
          LUKE\textsubscript{Large} & RoBERTa+EASA\textsuperscript{\textdagger} & \citet{yamada-etal-2020-luke} & \textit{RP} & till 2019 & +1.10 & 0 \\
          \rowcolor{Gray}
          & & & WP & till 12/2018 & & \\
          \hline
      \end{tabular}
  }
  \caption{Details about the models selected, sorted by $\Delta$ F\textsubscript{1}.
  We list the models' architectures and word embeddings, pre-training corpora,
  and the temporal coverage of the corpora.
  If the exact temporal coverage cannot be found, we report the time of publication of the corpus followed by *.
  For each model, we report the percentage change in F\textsubscript{1} and the change in ranking.
  Abbreviations: \textbf{BC} = BookCorpus \citep{Zhu2015AligningBA},
  \textbf{\textit{BP}} = BERT Pre-training Corpus,
  \textbf{CC} = CC-100 \citep{conneau-etal-2020-unsupervised},
  \textbf{CN} = CC-News \citep{nagel_2016},
  \textbf{C4} = Colossal Clean Crawled Corpus \citep{t5},
  \textbf{G5} = Gigaword5 \citep{gigaword},
  \textbf{OS} = OSCAR \citep{OrtizSuarezSagotRomary2019},
  \textbf{OW} = OpenWebText \citep{Gokaslan2019OpenWeb},
  \textbf{RN} = \textsc{RealNews} \citep{Zellers2019DefendingAN},
  \textbf{\textit{RP}} = RoBERTa Pre-training Corpus,
  \textbf{SS} = Semantic Scholar \citep{cohan-etal-2019-structural},
  \textbf{ST} = Stories \citep{stories},
  \textbf{WP} = Wikipedia,
  \textbf{1B} = 1B Benchmark \citep{Chelba2014OneBW}.
  \textsuperscript{\textdagger}Entity-aware self-attention \citep{yamada-etal-2020-luke}.
  }
  \label{tab:models}
\end{table*}

\noindent{\bf Annotation:}
We manually labeled this new dataset, which we refer to as CoNLL++, using the BRAT annotation interface \citep{stenetorp-etal-2012-brat}.
Articles were distributed between two authors, where one author annotated 96.1\% of the articles and the other annotated 50.0\%.
The first author's annotation is used as the gold standard.\footnote[3]{Articles only annotated by the second author were reviewed and then used as the gold standard.}
During the annotation process, articles from the CoNLL-2003 test set were interleaved with new articles from 2020,
in order to measure how closely the annotators follow the style of the original dataset.
%To ensure that CoNLL++ follows the original annotation guidelines and style, we interleaved the original CoNLL-2003 test set.

\noindent{\bf Inter-Rater Agreement:}
We find that the CoNLL++ annotations closely follow the style of the original dataset.
When considering labels in the CoNLL-2003 test set as gold,
our manual re-annotation achieves a 95.46 F\textsubscript{1} score.\footnote[4]{For reference, the current state of the art for automated NER taggers is 94.60 \citep{wang-etal-2021-automated}.}
The second author's annotation, when considering the first author's as gold,
receives a 96.23 F\textsubscript{1} score on overlapping articles.
The token-level Cohen's Kappa between the two authors is 97.42,
which can be considered almost perfect agreement \citep{artstein-poesio-2008-survey}.
Table~\ref{tab:stat} summarizes the statistics of the two datasets.

% Experimental Setup
\section{Experimental Setup}
\label{sec:experiment_setup}

We select models with a variety of architectures and pre-training corpora
and fine-tune these models to study how different factors affect generalization.
None of the models used any pre-training data that temporally overlap with CoNLL++, eliminating the possibility of articles
in CoNLL++ appearing in any pre-training corpus.
A list of all models and their implementation details can be found in Table~\ref{tab:models}.

Scripts for fine-tuning Flair and ELMo are adapted from \citet{reiss-etal-2020-identifying}.
\footnote[5]{Implementation from \citet{reiss_muthuraman_eichenberger_cutler_xu_2021}}
Other recurrent neural network (RNN) models are trained using various GitHub repositories (see footnotes 6, 7 and 8).
We fine-tune the BERT and RoBERTa models with the HuggingFace \verb+transformers+ library \citep{wolf-etal-2020-transformers},
except LUKE with AllenNLP \citep{gardner-etal-2018-allennlp}. T5 is fine-tuned to conditionally generate
NER tags around entities (e.g. <per> Jane Doe </per>).

% Table footnotes
\footnotetext[6]{Implementation from \citet{jie_2020}}
\footnotetext[7]{Implementation from \citet{kanakarajan_2019}}
\footnotetext[8]{Implementation from \citet{reimers-gurevych-2017-reporting}}

%\subsection{Hyperparameter Tuning}
%\label{sec:hpt}
A hyperparameter search is conducted for each model. 
%For the transformer-based models, we searched for the batch size and learning rate needed for the best model. 
We follow the recommended search space for a model if available in its publication. 
\{8, 16, 32\} and \{1e-5, 2e-5, 3e-5, 5e-5\} are used for most searches for batch sizes and learning rates
respectively. Appendix~\ref{sec:hpp_search} provides more details on the hyperparameter search.

%The hyperparameter search for the BiLSTM+GloVe models included another step of selecting between the SGD and Adam optimizer \citep{adam}.
%We adapted the code scripts for the Flair and ELMo models from \citet{Reiss2020IdentifyingIL} and as the performances were close to the numbers reported in their original publications, we did not do any hyperparameter tuning.
%Using the set of hyperparameters with the highest F$_1$ score on the CoNLL-2003 dev set, we train the models on the CoNLL-2003 train set for 10 epochs, and save the best epoch for evaluation. 
%Using the set of hyperparameters with the highest F$_1$ score on the CoNLL-2003 dev set, 
We train models on the CoNLL-2003 training set for 10 epochs, and use the dev set to select the best epoch and other hyperparameters for evaluation. 
Each model is evaluated five times with different random seeds on the CoNLL-2003 test set and on CoNLL++ to obtain the average F\textsubscript{1}.

In Table~\ref{tab:models}, we report the percentage change of F\textsubscript{1}, calculated as:
\begin{equation*}
  \Delta \text{F}\textsubscript{1} = \frac{\text{F}\textsubscript{1}^\text{CoNLL++} - \text{F}\textsubscript{1}^\text{CoNLL-2003}}{\text{F}\textsubscript{1}^\text{CoNLL-2003}} \times 100
\end{equation*}
where $\text{F}\textsubscript{1}^\text{CoNLL++}$ and $\text{F}\textsubscript{1}^\text{CoNLL-2003}$ are the F\textsubscript{1} scores on the CoNLL++ and CoNLL-2003 test sets respectively.
The results are visualized in Figure \ref{fig:results}. Raw F\textsubscript{1} scores are shown in Table~\ref{tab:detail_results} in the Appendix
(\S~\ref{ssec:app_orig_vs_new}).

% section: results
\section{What Ingredients are Needed for Good Generalization?}
As we can see in Figure \ref{fig:results} and Table \ref{tab:models},
different models have very different generalization.
Some models (e.g. RoBERTa-based models and T5$_\text{3B}$),
have no performance drop on CoNLL++, whereas other models' performances decrease significantly. 

%We define a model's generalization as the percentage change in F\textsubscript{1} score from CoNLL-2003 to CoNLL++, and 

In the following sub-sections, we evaluate the impact of a number of factors on generalization.  In \S \ref{sec:performance_drop}, we attempt to disentangle to what extent the observed performance drops on CoNLL++ are caused by temporal deterioration, or adaptive overfitting. 
%the cause of the performance drop of some models .

%the following factors on generalization: model size (\S~\ref{ssec:effect_of_size}),model architecture (\S~\ref{ssec:architecture})andnumber of training examples (\S~\ref{ssec:amount_of_training_examples}).
%We expect that each of the factors above plays an important role in achieving good generalization, specifically, we hypothesize that having a Transformer model architecture, larger size, pre-training corpus that is temporally closer to and in-domain with the test set, and more training examples in fine-tuning will lead to better generalization. We also expect to see some adaptive overfitting to the CoNLL-2003 test set.

%In the sections below, we analyze the effect of these factors to provide insights on the performance changes of the models, and further improving our understanding of building NER models that generalize well to new data.

% subsection: effect of size
\subsection{Model Size}
\label{ssec:effect_of_size}
It has been shown that the size of pre-trained models affects their performance \citep{scaling,t5}.
This inspired us to investigate the effect of model size on generalization.
We compare the performance of BERT, RoBERTa, ALBERT and T5 models with different
sizes on CoNLL++ and CoNLL-2003.  The results are visualized in Figure~\ref{fig:model_size}.
Details are available in Table~\ref{tab:model_size} in the Appendix (\S~\ref{ssec:app_model_size}).

We observe, from Table~\ref{tab:model_size}, that larger models perform better
on both test sets, but more importantly, as illustrated in Figure~\ref{fig:model_size}, performance degradation on CoNLL++ diminishes or even disappears as the model size grows.
The only exception is the RoBERTa-based models, whose base-sized model already achieves comparable performance on CoNLL++.
Figure~\ref{fig:model_size} suggests that larger model sizes not only increase performance on a static test set,
but also help models generalize better to new data.

\begin{figure}[h!]
  \centering
  \includegraphics[width=0.45\textwidth]{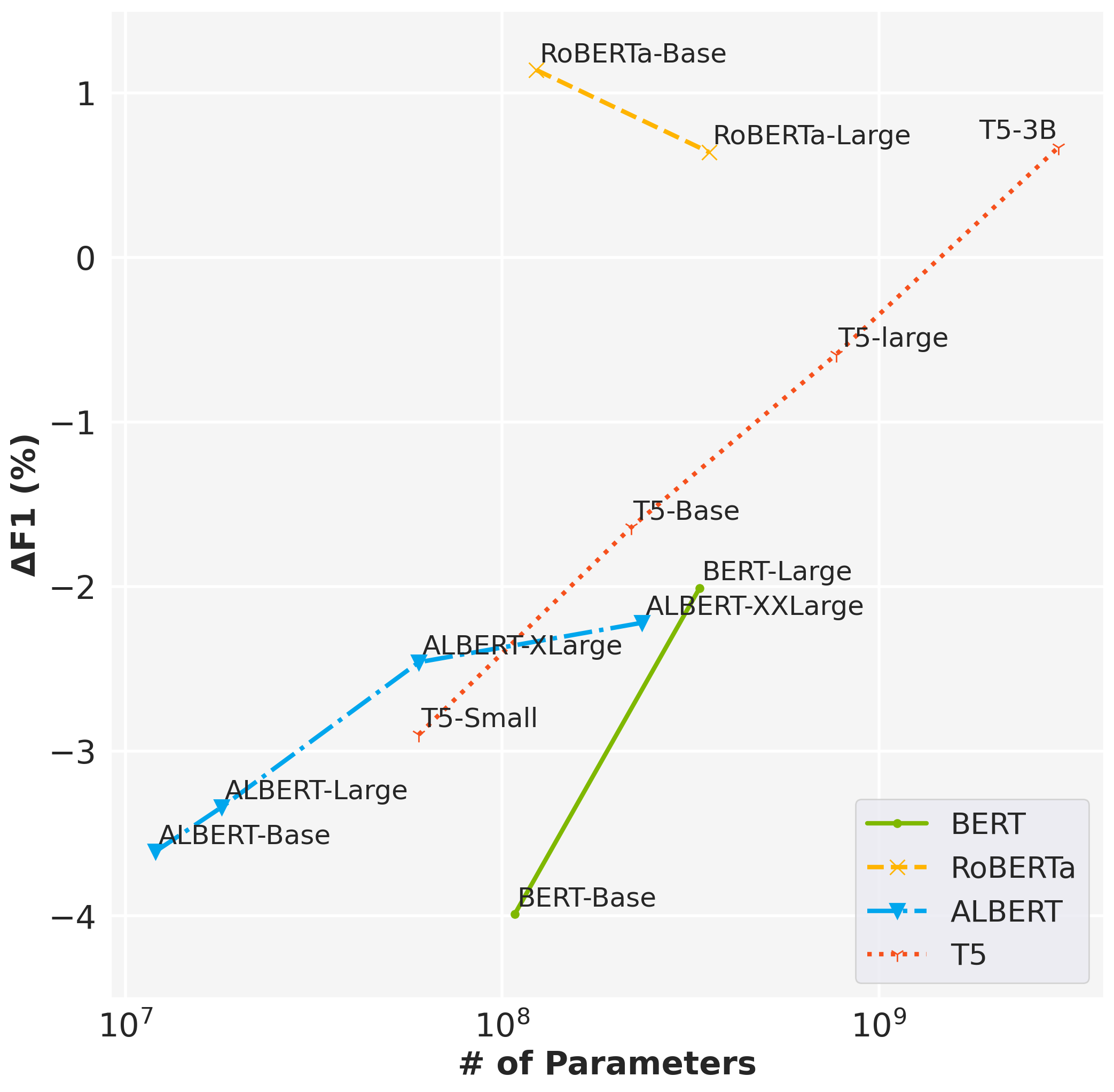}
  \caption{Plot of percentage change in F\textsubscript{1} scores ($\Delta$F\textsubscript{1}) against the number of parameters in log scale.
  All models except RoBERTa show an improvement in generalizability as the model size grows.
  }
  \label{fig:model_size}
\end{figure}

% Not all models grow linearly and different performance with the same number of parameters
It is also informative to look at the individual trend within each model family. Whereas T5 models
exhibit a linear relationship between the log number of parameters and $\Delta$F\textsubscript{1}, the improvement
of $\Delta$F\textsubscript{1} for ALBERT models diminishes as the size grows larger. Additionally,
models of similar sizes do not necessarily exhibit similar generalization.
For example,
BERT\textsubscript{Base} \& RoBERTa\textsubscript{Base} ($\sim$100M),
ALBERT\textsubscript{XXLarge} \& T5\textsubscript{Base} ($\sim$220M) and
BERT\textsubscript{Large} \& RoBERTa\textsubscript{Large} ($\sim$300M) 
all have similar sizes, but the performance changes within each pair are very different.

Both RoBERTa\textsubscript{Large} and T5\textsubscript{3B} achieve
a performance increase of $\sim$0.6\%, but the number of parameters of T5\textsubscript{3B} is $\sim$10 times of
that of RoBERTa\textsubscript{Large}. This shows that the generalizability of model is also affected by factors
other than the size of the model, but with the same architectures, larger models tend to generalize better.

\subsection{Model Architecture}
\label{ssec:architecture}
Based on the results from Table~\ref{tab:models}, we also observe that model architecture has a significant impact on generalizability.
Most BERT, RoBERTa and T5 models have a small performance drop (less than 4\% F\textsubscript{1}) on CoNLL++. 
The performances of RoBERTa\textsubscript{Large}, news-RoBERTa\textsubscript{Base}, LUKE and Longformer\textsubscript{Base}
improved slightly on CoNLL++. 
%Better generalization of the RoBERTa-based models can likely be attributed to the pre-training corpus being temporally closer to CoNLL++. 
The fact that most Transformer-based models achieve higher rankings in CoNLL++ also
confirms that pre-trained Transformers generalize better to new data.

BiLSTM models with Flair and ELMo embeddings, despite performing exceptionally on CoNLL-2003,
show larger performance drops on CoNLL++ (5-6\% F\textsubscript{1}), and the performance
of BiLSTM+GloVe models drops even more significantly (greater than 6\% F\textsubscript{1}). Such results
show a clear trend that Transformer-based models generalize better to new data.

\subsection{Number of Fine-Tuning Examples}
\label{ssec:amount_of_training_examples}
\begin{figure}[h!]
  \centering
  \includegraphics[width=0.45\textwidth]{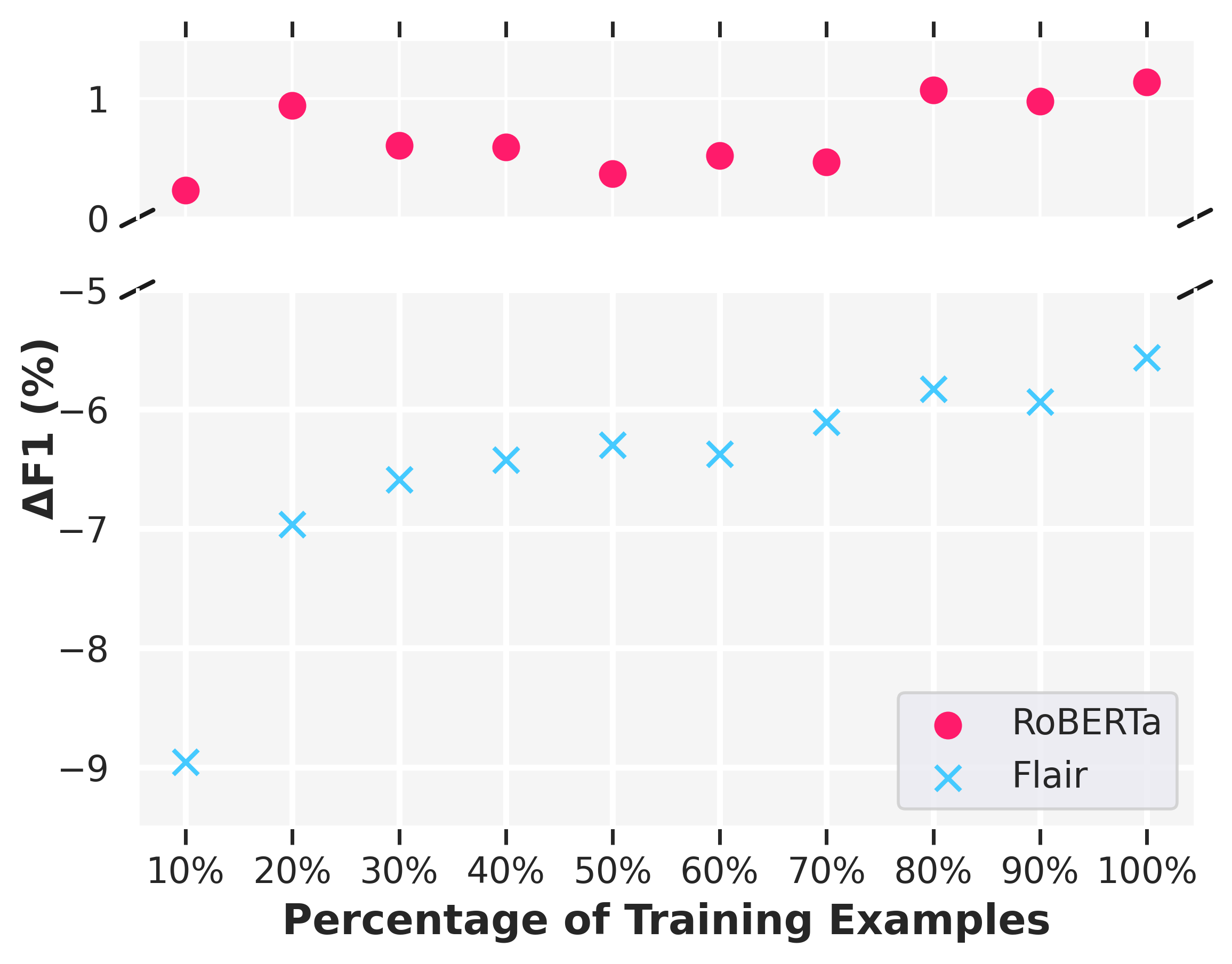}
  \caption{Plot of change in F\textsubscript{1} scores ($\Delta$F\textsubscript{1}) against the percentage of CoNLL-2003 training data used for fine-tuning.
  Both RoBERTa\textsubscript{Base} and Flair show improved generalization as we use more training examples,
  although Flair shows a more pronounced improvement.
  }
  \label{fig:training_examples}
\end{figure}

The generalizability of a model may also be affected by the size of the fine-tuning dataset.
We conduct experiments varying the number of CoNLL-2003 training examples used for fine-tuning from
10\% to 100\%. The fine-tuning is done with RoBERTa\textsubscript{Base}
and Flair embeddings using the same experimental setup as in Section~\ref{sec:experiment_setup}.
We plot the percentage change in F\textsubscript{1} against the percentage of training examples
in Fig~\ref{fig:training_examples}.

Both RoBERTa\textsubscript{Base} and Flair embeddings show improved generalization as
we use more training examples. However, this improvement is more pronounced for Flair than
RoBERTa\textsubscript{Base}. Even with 10\% of the training data, RoBERTa\textsubscript{Base}
already shows a positive change in F\textsubscript{1} scores, and increasing the amount of training
data to 100\% only improves the change by an absolute value of 1\%. In contrast, increasing the amount
of training data from 10\% to just 20\% can already improve $\Delta$F\textsubscript{1} by 2\% for Flair.

The empirical evidence supports our hypothesis that having more training examples can improve
the generalizability of the model, but such effect may vary across different models. 
RoBERTa-based models generalize well to new data even when only a small amount of fine-tuning data is available, whereas Flair benefits much more from having more fine-tuning data.
%Models already showing a strong generalizability such as RoBERTa\textsubscript{Base} may not benefit as much from more training examples.

% section: Performance drop
\section{What Causes the Performance Drop Observed for Some Models?}
\label{sec:performance_drop}
Models in Table~\ref{tab:models} show different levels of performance drop, or sometimes
performance gain, on CoNLL++ compared to the CoNLL-2003 test set,
and it is not entirely clear what causes this difference. We hypothesize
two potential causes, namely adaptive overfitting (\S~\ref{ssec:adaptive_overfitting})
and temporal drift (\S~\ref{ssec:temporal_drift}). In this section, we investigate each of these potential causes.

% subsection: adaptive overfitting
\subsection{Adaptive Overfitting}
\label{ssec:adaptive_overfitting}
% \ar{I feel like this section can still be improved.  E.g. we probably want to make it a bit clearer exactly what we can conclude from Figure 2.  I think the slope of the line tells us something about the overall generalization of models, but there could still be some individual models that are overfitting. 
%  We need to be a bit clearer about what is the logic of the argument we are making here, and also make sure our claims are scoped appropriately.}
% \ar{Add a bit of background to teach the reader about diminishing returns + define adaptive overfitting.  Also define test reuse.}

We first investigate if the performance drop is caused by adaptive overfitting of models
developed over the past 20 years. \citet{roelofs2019meta}
defined adaptive overfitting as the overfitting caused by reusing the same test set (\textit{test reuse}).
\citet{recht2019imagenet} studied this phenomenon in the context of ImageNet by measuring
to what extent can the improvement on the old test set translate to improvement on the new test set (\textit{diminishing return}).
We analyze both effects to conclude the presence of adaptive overfitting.

% subsubsection: diminishing return
\subsubsection{Diminishing Return}
\label{sssec:diminishing_return}
Following \citet{recht2019imagenet}, we measure the diminishing return on the CoNLL++ test set.
Diminishing return measures if the improvement on CoNLL-2003 test set, gained by the continuous
effort of developing NER taggers over 20 years, translates to smaller (\textit{diminishing})
improvement on CoNLL++.

We fit a line to the data points in Figure~\ref{fig:results}, and then calculate its slope.
A slope greater than 1 indicates that every unit of improvement on the CoNLL-2003 test set by the
development of models translates to more than one unit of improvement on CoNLL++,
i.e. there is no diminishing return.
% The equation of the best-fit line is \[
%   \text{F\textsubscript{1}\textsuperscript{new}} = -162.03 + 2.729 \cdot \text{F\textsubscript{1}\textsuperscript{orig}}
% \]
% where F\textsubscript{1}\textsuperscript{new} and F\textsubscript{1}\textsuperscript{orig} are the
% F\textsubscript{1} scores on the CoNLL++ and CoNLL-2003 test sets respectively.
% The slope 2.729 
We measure the slope to be 2.729 $>$ 1, indicating that we have not found any diminishing return on CoNLL++, and therefore
\textit{no} adaptive overfitting caused by the model development over the past two decades.
% \ar{This is a bit of a strong claim "no adaptive over-fitting", as some individual models certianly look like they are over-fit (e.g. Flair and BiLSTM-CRF.}

% subsubsection: test reuse
\subsubsection{Test Reuse}
\label{sssec:test_reuse}

% \ar{This experiment seems to be fairly important for the overall story of the paper.
% I think it is presenting fairly conclusive evidence that most of the performance
% difference is due to temporal drift, and not adaptive overfitting.
% Need to make this argument clearer.
% What are the claims we are making in this section, and how does the experiment
% support them?
% }
If the models are overfitting to the CoNLL-2003 test set due to test reuse, we should see not only
a performance degradation on CoNLL++, but also a performance degradation on a test set
taken from the same distribution as the CoNLL-2003 test set.

To obtain a new test set taken from the same distribution as the CoNLL-2003 test set, we
resampled a new train/dev/test split from the CoNLL-2003 dataset, which we call CoNLL-2003'.
Each split contains the same number of articles
% (marked by ``DOCSTART")
as its corresponding split in the CoNLL-2003 dataset.
The ``new'' test set is thus certain to come from the same distribution as the original CoNLL-2003 test.
We train and evaluate models on CoNLL-2003' with the same experimental setup as
in Section~\ref{sec:experiment_setup}, and report the results in Table~\ref{tab:test_reuse}.

\begin{table}[ht!]
  \resizebox{\columnwidth}{!}{
      \centering
      \tiny
      \begin{tabular}{|l|c|c|}
          \hline
          \textbf{Name} & \begin{tabular}{@{}c@{}} \textbf{CoNLL++} \\ \textbf{$\Delta$F\textsubscript{1} (\%)} \end{tabular} & \begin{tabular}{@{}c@{}} \textbf{CoNLL-2003'} \\ \textbf{$\Delta$F\textsubscript{1} (\%)} \end{tabular} \\
          \hline
          \rowcolor{Gray}
          BiLSTM-CRF & -20.25 & +2.53 \\
          BiLSTM-CNN & -15.09 & +1.75 \\
          \rowcolor{Gray}
          SciBERT & -8.94 & -0.09 \\
          BiLSTM-CNN-CRF & -6.52 & +2.95 \\
          \rowcolor{Gray}
          % SpanBERT\textsubscript{Large} & 88.31\textsubscript{0.41} & 88.49\textsubscript{0.86} & +0.20 \\
          BiLSTM-CRF-ELMo & -5.72 & +1.58 \\
          Flair & -5.57 & +0.76 \\
          \rowcolor{Gray}
          Pooled Flair & -4.65 & +1.60 \\
          mBERT & -4.22 & +2.80 \\
          \rowcolor{Gray}
          GigaBERT & -3.90 & +1.75 \\
          ALBERT\textsubscript{Base} & -3.61 & +2.36 \\
          \rowcolor{Gray}
          BERT\textsubscript{Large} & -2.01 & +0.47 \\
          XLM-RoBERTa\textsubscript{Base} & -0.90 & -1.52 \\
          \rowcolor{Gray}
          T5\textsubscript{Large} & -0.59 & +2.65 \\
          RoBERTa\textsubscript{Large} & +0.64 & +0.38 \\
          \rowcolor{Gray}
          Longformer\textsubscript{Base} & +1.00 & +2.44 \\
          news-RoBERTa\textsubscript{Base} & +1.06 & +1.90 \\
          \rowcolor{Gray}
          Luke & +1.10 & +1.87 \\
          \hline
      \end{tabular}
  }
  \caption{Comparison between the performance change on CoNLL++ and CoNLL-2003' test sets.
  The table shows clearly that the performances of most models are not degrading because of test reuse.
  Detailed results can be found in Table~\ref{tab:test_reuse_stats} in Appendix~\ref{ssec:app_test_reuse}.
  }
  \label{tab:test_reuse}
\end{table}

We only observe SciBERT and XLM-RoBERTa\textsubscript{Large} models performing slightly worse on the CoNLL-2003' test set, while
all other models appear to perform better. Most models suffering from performance degradation
on the CoNLL++ also perform better on the CoNLL-2003' test set. This provides
evidence that individual models are \textit{not} overfitting to the CoNLL-2003 test set.

Based on our results above, the performance degradation on the CoNLL++ is likely \textit{not} caused by overfitting on CoNLL-2003.
Rather, it is more likely caused by temporal drift, which we discuss in the next section.

% subsection: temporal drfit
\subsection{Temporal Drift}
\label{ssec:temporal_drift}
% \ar{The following two paragraphs need to be re-written from scratch for clearly.  This needs to be very specific about what the prior work did, and how your analysis is different. 
%  As written, the reader will be confused about this.  For example, it is not clear what the following phrases are supposed to mean: (1) ``showing such effect on GloVe and Flair embeddings.'', (2) ``reported temporal deterioration on GloVe and not RoBERTa, but later mentioned that
% the temporal overlap between the RoBERTa pre-training corpora of the embeddings and the downstream task posed a challenge to
% isolate the effect of pre-training corpora.''}
% Previous studies have shown that the performance on NER is affected by how close the test set is to the
% train set, such as \citet{rijhwani-preotiuc-pietro-2020-temporally} showing such effect on GloVe and Flair embeddings.
% \citet{agarwal-nenkova-2022-temporal} reported temporal deterioration on GloVe and not RoBERTa, but later mentioned that
% the temporal overlap between the RoBERTa pre-training corpora of the embeddings and the downstream task posed a challenge to
% isolate the effect of pre-training corpora.

% In this section, we investigate how the generalization of models is affected by
% the temporal difference between the pre-training corpus of the model and the test set.

\textit{Temporal drift} refers to the performance degradation of a model on the
downstream task caused by the temporal difference between the train and test data.
Prior work has shown that the performance on NER is affected by temporal drift.
For example, \citet{rijhwani-preotiuc-pietro-2020-temporally} showed that the performance
of GloVe and Flair embeddings on NER degrades when the test data is more temporally
distant from the train data of the downstream task. \citet{agarwal-nenkova-2022-temporal} also
reported the same observation on GloVe embeddings.

In this section, we use the same term ``temporal drift'' but refer to the
deterioration of \textit{generalization} of models caused by the temporal
difference between the pre-training corpus of their word embeddings and the test data
of the downstream task. We hypothesize that generalization is largely affected
by such temporal drift. We conduct experiments on Flair and ELMo, as well as on RoBERTa.

% subsection: temporal drift in CWE
%\subsubsection{Temporal Drift in Contextualized Word Embeddings}
\subsubsection{Temporal Drift in Flair and ELMo}
\label{ssec:td_cwe}
We first investigate if bringing the pre-training corpora of Flair and ELMo closer to the test set
can improve their generalizability. We notice that both embeddings were trained on 
1B Benchmark \citep{Chelba2014OneBW}. This corpus was collected from WMT11 \citep{wmt11} English monolingual data,
which is largely comprised of news data between 2007-2011. We hypothesize that pre-training these
embeddings on a more recent corpus, e.g. \textsc{RealNews} corpus \citep{Zellers2019DefendingAN}
which contains news articles from 2016-2019, will improve their generalizability.

% Both CoNLL-2003 and CoNLL++ are comprised of news articles. Based on Table~\ref{tab:detail_results}, we observe that models pre-trained with news data,
% such as all RoBERTa-based models, generalize to CoNLL++ better. Longformer\textsubscript{Base}
% and news-RoBERTa\textsubscript{Base} were further pre-trained on the \textsc{RealNews} corpus,
% and they demonstrate larger improvements on CoNLL++ compared to RoBERTa\textsubscript{Large}.
% SciBERT and mBERT were pre-trained on out-of-domain data, and exhibit worse generalization.

% However, it is not clear whether such domain-dependence is present in the LSTM-based models.
% To this end, we trained BiLSTM models with Flair and ELMo embeddings on the \textsc{RealNews} corpus \citep{Zellers2019DefendingAN}
% and evaluated them on CoNLL++.  The results are shown in Table \ref{tab:in_domain}.

\begin{table}[h!]
  \resizebox{\columnwidth}{!}{
    % \centering
    % \tiny
    \begin{tabular}{|l|c|c|c|}
      \hline
      \textbf{Name} & \textbf{CoNLL-2003} & \textbf{CoNLL++} & \textbf{$\Delta$F\textsubscript{1} (\%)} \\
      \hline
      Flair & 92.46\textsubscript{0.14} & 87.31\textsubscript{0.69} & -5.57 \\
      \rowcolor{Gray}
      Flair\textsubscript{RN} & 90.91\textsubscript{0.22} & 88.46\textsubscript{0.69} & \textbf{-2.69} \\
      \hline
      Pooled Flair & 93.15\textsubscript{0.24} & 88.82\textsubscript{0.60} & -4.65 \\
      \rowcolor{Gray}
      Pooled Flair\textsubscript{RN} & 92.98\textsubscript{0.14} & 89.73\textsubscript{0.27} & \textbf{-3.50} \\
      \hline
      ELMo & 92.36\textsubscript{0.10} & 87.08\textsubscript{0.39} & -5.72 \\
      \rowcolor{Gray}
      ELMo\textsubscript{RN} & 92.11\textsubscript{0.07} & 90.79\textsubscript{0.50} & \textbf{-1.43} \\
      \hline
    \end{tabular}
  }
  \caption{Percentage change in F\textsubscript{1} scores ($\Delta$F\textsubscript{1})
  on CoNLL++ of Flair and ELMo embeddings when pre-trained on 1B Benchmark vs on
  \textsc{RealNews} corpus. Pre-training on \textsc{RealNews}, which is temporally
  closer to CoNLL++, improves the generalization of Flair and ELMo embeddings.}
  \label{tab:td_cwe}
\end{table}

% \textcolor{red}{(How to differentiate b/w TD and ID? Move to temporal section and mention in-domain-ness there?)}
% Both embeddings were intially trained on the 1B Benchmark \citep{Chelba2014OneBW},
% which was collected from WMT11 \citep{wmt11}
% including the Europarl corpus \citep{koehn-2005-europarl}, news commentary, and
% news data between 2007-2011.
To control the experiment, we randomly sample 1 billion tokens of data from \textsc{RealNews}.
We train Flair embeddings following the
same procedure detailed in \citet{akbik-etal-2018-contextual} to obtain both the forward and backward embeddings.
Our embeddings achieve character level perplexity on the test set of 2.45 for the forward embeddings and 2.46 for the backward embeddings,
comparable to 2.42 reported in \citet{akbik-etal-2018-contextual}.
Similarly, we train ELMo embeddings following \citet{peters-etal-2018-deep}, which achieves a
perplexity of 40.07 on the test set, comparable to 39.7 reported. 
We use the same training scripts and hyperparameters as our experiments in Section~\ref{ssec:architecture}
for Flair, Pooled Flair and ELMo. The newly trained models are dubbed as Flair\textsubscript{RN},
Pooled Flair\textsubscript{RN} and ELMo\textsubscript{RN}.

% It is clearly shown in Table \ref{tab:in_domain} that pre-training on in-domain data improves generalization to new data,
% reflected by the smaller performance degradation between the CoNLL-2003 and CoNLL++ test sets.
It is clearly shown in Table~\ref{tab:td_cwe} that having the training corpus for Flair and
ELMo embeddings temporally closer to the CoNLL++ test set improves generalization.
Notably, the performance gap for ELMo is reduced to -1.43\%, better than that of
BERT\textsubscript{Large} (-2.01\%). The improvements in generalization are attributed to
the performance drops on the CoNLL-2003 test set and improvements on CoNLL++.
% i.e. temporal drift mentioned in Rijhwani et al. and Agarwal et al.

% This observation suggests that the generalizability of the LSTM-based
% contextualized word embeddings is affected by the domain of the pre-training data. However, even with
% in-domain data, the LSTM-based models still suffer from performance drops on CoNLL++ as compared to the CoNLL-2003 test set,
% which suggests these models do not generalize as well as the transformer-based models.
This provides evidence that the generalizability of the LSTM-based contextualized word embeddings
is affected by temporal drift. However, even temporally closer data, these models still
suffer from performance drops. This suggests that other ingredients, such as
model architecture (\S~\ref{ssec:architecture}), are still needed for a good generalization.

\subsubsection{Temporal Drift in RoBERTa}

Because pre-training a transformer model from scratch is expensive, we continue pre-training
from the RoBERTa\textsubscript{Base} checkpoint, leveraging the findings from \citet{gururangan-etal-2020-dont}
that models learn to adapt to the distribution of the new corpora with continued pre-training.

We use the WMT20 English dataset \citep{barrault-etal-2019-findings}, consisting of English news data
from 2007 to 2021. To avoid temporal overlap, we only use data from 2007 to 2019 as the pre-training
corpora. The data are divided by year, and we preprocess the data such that the number of tokens per
year is the same. We train the RoBERTa\textsubscript{Base} model for 3 epochs with the masked language
modeling (MLM) objective. Checkpoints from each year are then fine-tuned on the CoNLL-2003 dataset,
with the same experimental setup described in Section~\ref{sec:experiment_setup}. We evaluate the models on the CoNLL-2003
test set and CoNLL++, and plot the results in Figure~\ref{fig:temporal_drift}.
Detailed performances are reported in Table~\ref{tab:temporal_drift} in the Appendix (\S~\ref{ssec:app_temporal_drift}).

\begin{figure}[h!]
  \centering
  \includegraphics[width=0.45\textwidth]{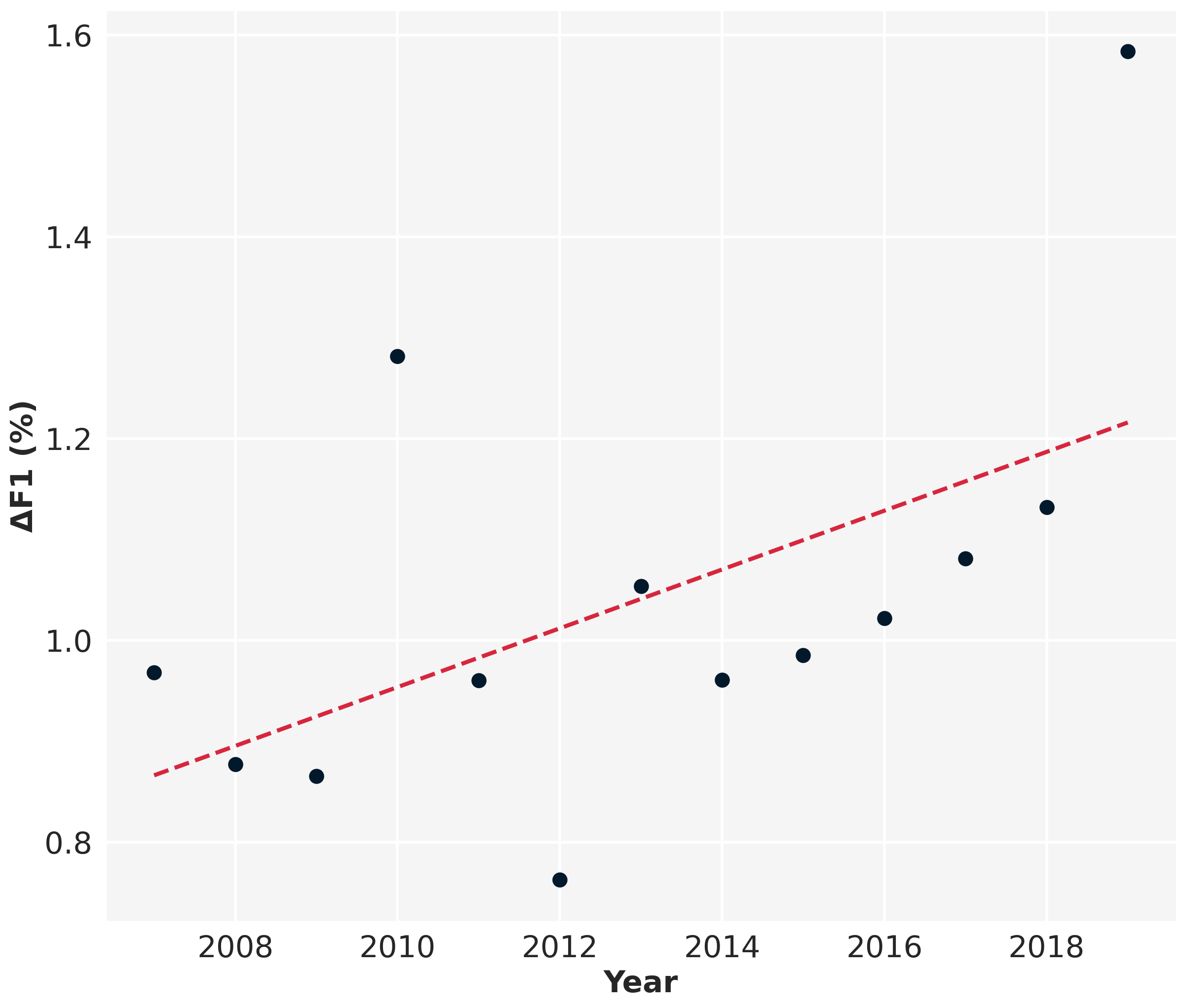}
  \caption{Plot of $\Delta$F\textsubscript{1} scores against the year of data used for RoBERTa pre-training.
  The upward trend, indicated by the dashed best-fit line, shows that the generalization improves
  as the pre-training corpora used is temporally closer to CoNLL++.
  }
  \label{fig:temporal_drift}
\end{figure}

The results show a clear trend of performance degradation when the pre-training corpora is temporally distant from
the test set. When the pre-training corpus is more recent, it becomes temporally closer to CoNLL++,
leading to better CoNLL++ performance, and hence better generalization. In Figure~\ref{fig:temporal_drift},
the $\Delta$F\textsubscript{1} shows an upward trend with a correlation coefficients of 0.55,
indicating a moderate positive correlation between generalization and the year of the pre-training corpora.
This suggests that generalization is affected by the effect of temporal drift.
This explains the better generalizability of models
such as LUKE\textsubscript{Large} and T5\textsubscript{3B},
pre-trained on temporally closer data to the CoNLL++ test set,
showing that temporal drift is the main driving factor for the different levels of generalization.

% Section: Related Work
\section{Related Work}

How well pre-trained LMs adapt to data from future time periods has undergone extensive study. Temporal degradation has been found to be
a challenge for many tasks, including language modeling \citep{lazaridou2021mind}, NER
\citep{augenstein2017generalisation,agarwal-nenkova-2022-temporal,rijhwani-preotiuc-pietro-2020-temporally,ushio-etal-2022-named}, QA \citep{dhingra-etal-2022-time},
entity linking \citep{zaporojets2022tempel}, and others \citep{luu-etal-2022-time,amba2021dynamic}. All of
this work has found that the performance of LMs degrades as the temporal distance between the
training data and the test data increases, sometimes called ``temporal misalignment'' \citep{luu-etal-2022-time}.
In contrast to the prior work, we study performance deterioration on a dataset that has been heavily used to develop NER models over a period of 20 years, and conduct extensive experiments that aim to disentangle the effects of aging training sets from those due to heavy test reuse.
%Our work adds to this line of research by studying how different language models developed over the years
%can generalize to a new NER test set when trained on data from twenty years ago. 

Most closely related to our work is \citet{agarwal-nenkova-2022-temporal}, who analyzed a recently created Twitter NER dataset
\citep{rijhwani-preotiuc-pietro-2020-temporally} over the period 2014-2019, and found no performance
deterioration when using RoBERTa-based representations. We build on this line of work by carefully
measuring performance deterioration of models trained on the CoNLL-2003 dataset when evaluated
on modern data. We analyze which factors are necessary for an NER model trained on a 20-year-old dataset
to generalize well to modern data.
Furthermore, the large 20-year gap helps us focus on not only temporal deterioration, but
also if the extensive test reuse leads to adaptive overfitting. We present evidence in support of the hypothesis
that most performance degradation is due to temporal drift and not adaptive overfitting.

Prior work has attempted to mitigate temporal degradation, mostly through continuously updating
LMs with new data \citep{jang-etal-2022-temporalwiki,jin-etal-2022-lifelong,loureiro-etal-2022-timelms}. \citet{luu-etal-2022-time} explored this
idea but found that temporal adaptation is not as effective as fine-tuning on the data from whose time period
the dataset is drawn. In addition, catastrophic forgetting \citep{Robins1995CatastrophicFR} can also be a problem when updating the LMs.
\citet{jin-etal-2022-lifelong} found that applying knowledge distillation \citep{hinton2015distilling} based approaches
to continual learning can mitigate catastrophic forgetting, while improving the temporal generalization of LMs.
\citet{dhingra-etal-2022-time} proposed to train the LMs with an additional temporal objective by conditioning
on the year of data, and found that this effectively mitigated catastrophic forgetting. \citet{jang-etal-2022-temporalwiki}
created a lifelong benchmark for continuous training and evaluating LMs. 

\begin{comment}
\textbf{Adaptive Overfitting} $\quad$ Our work was inspired by \citet{recht2019imagenet} and \citet{roelofs2019meta},
which discussed the possible adaptive overfitting to test sets on image recognition tasks and found little evidence.
\citet{recht2019imagenet} provided a way to measure adaptive overfitting by observing if there is
any diminishing returns on the new dataset. Our work extends this idea to NER tasks and finds
no empirical evidence of adaptive overfitting. 
\end{comment}

\section{Conclusion and Future Directons}
% \ar{The conclusion can be improved.  Can paraphrase the intro/abstract.}
In this paper, we evaluate the generalization of NER models using CoNLL++,
a CoNLL-style annotated NER test dataset with data from 2020.
We conduct experiments on more than 20 models and find that models exhibit different
generalizability.

Surprisingly, we find that generalizability is \textit{not} affected by
adaptive overfitting, but rather by temporal drift.
To achieve better generalization, we need the combination of four factors:
a modern transformer-based architecture,
a large number of parameters, 
a large amount of fine-tuning data
and a temporally closer pre-training corpus to the test set.
We find that our progress on developping NER taggers is largely successful, showing not only
good performance on individual test set but also good generalization on new data.
This allows CoNLL-2003 taggers to still work in 2023.

Future research can focus on ways to mitigate temporal drift.
Investigation on attributes of pre-training or fine-tuning
corpora that causes temporal drift, such as change of
entities mentioned, different usage of language, etc., can
also shed light on the more specific impacts from temporal drift, thereby
inspiring new and better ways to mitigate it.

We hope that our work provides insights on factors affecting generalization and how to
mitigate the negative impact, and calls for more research on this everlasting problem of
generalization in the NLP community.
% Our results show that a good generalization is
% achieved by having a large transformer-based model, pre-trained on a corpus that is
% in-domain with and temporally closer to the test set, and fine-tuned on a large amount of data. Our analysis on adaptive overfitting
% suggests that researchers have been developing recent models that not only perform well on the CoNLL-2003
% test set, but also generalize better to new and unseen data. The presence of models with a
% good performance on the CoNLL-2003 test set but poor generalization suggests the value of
% annotating more datasets to study how well NLP models generalize to modern data in the future.

\section{Limitations}
% Our experiments do not attempt to distangle the effect of adaptive overfitting from temporal drift.\footnote[8]{We did not attempt to annotate a new test set from the same time period as the original CoNLL test set, which we leave for future work.  Note that controlling all possible factors to build a test set from a nearly identical distribution is known to be a challenging task \citep{recht2019imagenet}.}
%Furthermore The more important question is: how well NER models will work on new data.
% However, it seems likely most of the performance degradation observed in our study is due to temporal drift rather than adaptive overfitting.  For example flair and ELMo embeddings' performance drop might be explained by the fact they were mostly pre-trained using data from 2011 and earlier. Recent work has also found little evidence for adaptive overfitting when using public test sets \citep{roelofs2019meta}.
Our analysis on temporal drift (\S~\ref{ssec:temporal_drift}) was limited by the fact that the developer of many models in
our study did not release the exact time period of the pre-training corpora used. Additionally,
models such as BERT and RoBERTa were pre-trained on corpora that could be potentially be
temporally close to the CoNLL++ test set.

In the section on test reuse (\S~\ref{sssec:test_reuse}), due to a limited compute budget,
we were only able to conduct this experiment on a single new train/dev/test split,
so it is possible that the new split happens to be easier than the mean of the
distribution. However, our experiments still provide additional evidence models are not
overfitting the original CoNLL-2003 test set.

It is worth noting that when using older models trained on the CoNLL-2003 dataset,
one additional reason for the performance degradation, especially in real-world
deployment, is that the data used to evaluate the models can be out-of-domain.
In our experiments, we attemped to control the domain of the test data on which the models were
evaluated to assess other factors for performance degradation. However, we acknowledge that
in reality, model performance can be affected by factors such as
the emerging text types (e.g. Twitter did not exist when CoNLL-2003
NER task was created), which leads to changes
in domain, and therefore affects the generalizability of the models.

We acknowledge that having CoNLL++ will not resolve the problem of generalization to modern data.
As new data keep emerging, there will always be the question of how well NER models generalize to
that new data. We hope that our paper will encourage researchers in the NLP community to continuously
annotate new test set to study this problem, so that we ensure the robustness and generalizability of models.

\section*{Acknowledgements}
We would like to thank Fan Bai, Yang Chen, Chao Jiang, Junmo Kang
and Mounica Maddela for providing feedback on earlier drafts of this paper,
We would also like to appreciate the comments from the
anonymous reviewers on how to improve the paper.
This material is based upon work supported by the NSF (IIS-2052498) and IARPA via the BETTER program (2019-19051600004). The views, opinions, and/or findings expressed are those of the author(s) and should not be interpreted as representing the official views or policies of the Department of Defense, IARPA or the U.S. Government.

\bibliography{anthology,custom}
% \bibliography{custom}
\bibliographystyle{acl_natbib}

%\clearpage
\newpage
\appendix

\section*{Appendix}
% \ar{need to fix formatting issues in the appendix}
\label{sec:appendix}

\section{Tabular Data in CoNLL-2003}
\label{sec:app_tabular_data}
We found a significant amount of documents in the CoNLL-2003 test set that list the outcomes of various sports events, which contributes to the larger
proportion of named entities in Table \ref{tab:stat}. These documents appear as though they may have been intended for display on news
tickers.\footnote[9]{\url{https://en.wikipedia.org/wiki/News_ticker}}  We present an example below.

\begin{tcolorbox}[breakable]
  SOCCER SHOWCASE-BETTING ON REAL MADRID V BARCELONA .

  MADRID 1996-12-06

  William Hill betting on Saturday 's

  Spanish first division match between Real Madrid and Barcelona :

  To win : 6-5 Real Madrid ; 7-4 Barcelona

  Draw : 9-4

  Correct score :

  Real Madrid to win Barcelona to win

  1-0 13-2 1-0 15-2

  2-0 9-1 2-0 12-1

  2-1 8-1 2-1 10-1

  3-0 20-1 3-0 28-1

  3-1 16-1 3-1 22-1

  3-2 25-1 3-2 25-1

  4-0 50-1 4-0 100-1

  4-1 40-1 4-1 80-1

  4-2 50-1 4-2 80-1

  Draw :

  0-0 8-1

  1-1 11-2

  2-2 14-1

  3-3 50-1

  Double result :

  half-time full-time

  5-2 Real Madrid Real Madrid

  14-1 Real Madrid Draw

  28-1 Real Madrid Barcelona

  5-1 Draw Real Madrid

  4-1 Draw Draw

  11-2 Draw Barcelona

  25-1 Barcelona Real Madrid

  14-1 Barcelona Draw

  4-1 Barcelona Barcelona

  First goalscorer of match :

  Real Madrid Barcelona

  9-2 Davor Suker 9-2 Ronaldo

  5-1 Pedrag Mijatovic 7-1 Luis Figo

  7-1 Raul Gonzalez 7-1 Juan Pizzi

  12-1 Fernando Redondo 9-1 Giovanni

  14-1 Victor Sanchez 12-1 Guillermo

  Amor

  16-1 Jose Amavisca 14-1 Roger Garcia

  16-1 Manolo Sanchis 14-1 Gheorghe

  Popescu

  16-1 Roberto Carlos 16-1

  JosepGuardiola

  20-1 Fernando Hierro 20-1 Ivan de

  laPena

  20-1 Luis Milla 25-1 Luis

  Enrique

  33-1 Fernando Sanz 25-1

  AbelardoFernandez

  40-1 Carlos Secretario 28-1 Sergi Barjuan

  40-1 Rafael Alkorta 33-1 Albert

  Ferrer

  40-1 Chendo Porlan 33-1 Miguel Nadal

  40-1

  Laurent Blanc
\end{tcolorbox}
Being one of the 231 articles (0.43\%) in the CoNLL-2003 test set, this article contains 59 (1.04\%) named entities,
including 23 ORG (1.38\%), 34 PER (2.10\%), 1 LOC and 1 MISC. We counted that there are in total 71 (30.7\%) such files
which contribute to 872 ORG (52.5\%), 889 PER (55.0\%), 657 LOC (39.4\%) and 159 MISC (22.6\%).

Additionally, as each line is considered to be a sentence in CoNLL-2003 dataset (separated by an empty line
in the original format), and as items by spaces are considered to be tokens, this also demonstrates why
the average token per sentence is much lower in CoNLL-2003 than in CoNLL++. The tabular data contains
much shorter sentences in plethora, which significantly lowers the average token per sentence.

\section{Hyperparameter Search}
\label{sec:hpp_search}
In this section, we include the details on how we conducted the hyperparameter search for the transformer-based models. We trained most models with different sets of hyperparameters for 10 epochs and save the checkpoints that achieved the highest dev F$_1$ score. For each model, we compare performance on the dev set of checkpoints trained with different hyperparameters and select the set of hyperparameters with the best performance.

We tuned the learning rate and batch sizes for all models. If the instructions on how to tune the hyperparameters for a model are stated in its publication, we followed the instructions as closely as possible. Otherwise, we would tune the model using a default set of hyperparameters, where learning\_rate = \{1e-5, 2e-5, 3e-5, 5e-5\} and batch\_size = \{8, 16, 32\}. Here we only list models for which we did not use the default set of hyperparameters.

\begin{itemize}
    \item ALBERT:
    \begin{itemize}
        \item learning\_rate = \{1e-5, 2e-5, 3e-5, 5e-5\}
        \item batch\_size = \{16, 32, 48, 128\}
    \end{itemize}
    \item GigaBERT:
    \begin{itemize}
        \item learning\_rate = \{1e-5, 2e-5, 5e-5, 1e-4\}
        \item batch\_size = \{4, 8, 16, 32\}
    \end{itemize}
    \item Longformer:
    \begin{itemize}
        \item learning\_rate = \{1e-5, 2e-5, 3e-5, 5e-5\}
        \item batch\_size = \{16, 32\}
        \item total\_num\_epoch = 15
    \end{itemize}
    \item news\_roberta\_base:
    \begin{itemize}
        \item learning\_rate = \{1e-5, 2e-5, 3e-5\}
        \item batch\_size = \{16, 32\}
    \end{itemize}
    \item XLM-RoBERTa:
    \begin{itemize}
        \item learning\_rate = \{1e-5, 2e-5, 3e-5, 5e-5\}
        \item batch\_size = \{16, 32\}
    \end{itemize}
    \item T5:
    \begin{itemize}
        \item learning\_rate = \{2e-5, 3e-5, 5e-5, 1e-4\}
        \item batch\_size = \{4, 8\}
    \end{itemize}
\end{itemize}

\section{Detailed Results}
\label{sec:detail_results}

In this section, we include all the performance statistics.

\subsection{CoNLL-2003 vs CoNLL++}
\label{ssec:app_orig_vs_new}
Table~\ref{tab:detail_results} shows the performance statistics of all
models on the CoNLL++ and CoNLL-2003 test sets.

\begin{table}[h!]
  \resizebox{\columnwidth}{!}{
      \centering
      \tiny
      \begin{tabular}{|l|c|c|c|}
          \hline
          \textbf{Name} & \textbf{CoNLL-2003} & \textbf{CoNLL++} & \textbf{$\Delta$F\textsubscript{1} (\%)} \\
          \hline
          \rowcolor{Gray}
          BiLSTM-CRF & 91.00\textsubscript{0.18} & 72.57\textsubscript{2.78} & -20.25 \\
          BiLSTM-CNN & 89.02\textsubscript{0.09} & 75.59\textsubscript{0.66} & -15.09 \\
          \rowcolor{Gray}
          Stanford NLP & 88.28 & 76.58 & -13.25 \\
          SciBERT & 87.05\textsubscript{0.91} & 79.27\textsubscript{1.43} & -8.94 \\
          \rowcolor{Gray}
          BiLSTM-CNN-CRF & 90.25\textsubscript{0.22} & 84.37\textsubscript{0.49} & -6.52 \\
          % SpanBERT\textsubscript{Large} & 88.31\textsubscript{0.41} & 84.06\textsubscript{0.54} & -6.52 \\
          BiLSTM-CRF-ELMo & 92.36\textsubscript{0.10} & 87.08\textsubscript{0.39} & -5.72 \\
          \rowcolor{Gray}
          Flair & 92.46\textsubscript{0.14} & 87.31\textsubscript{0.69} & -5.57 \\
          Stanza & 91.50 & 86.75 & -5.19 \\
          \rowcolor{Gray}
          Pooled Flair & 93.15\textsubscript{0.24} & 88.82\textsubscript{0.60} & -4.65 \\
          mBERT & 91.06\textsubscript{0.42} & 87.22\textsubscript{0.56} & -4.22 \\
          \rowcolor{Gray}
          GigaBERT & 91.35\textsubscript{0.27} & 87.79\textsubscript{0.37} & -3.90 \\
          ALBERT\textsubscript{Base} & 89.53\textsubscript{0.23} & 86.30\textsubscript{0.39} & -3.61 \\
          \rowcolor{Gray}
          BERT\textsubscript{Large} & 91.77\textsubscript{0.20} & 89.93\textsubscript{0.74} & -2.01 \\
          XLM-RoBERTa\textsubscript{Base} & 91.04\textsubscript{0.53} & 90.22\textsubscript{0.77} & -0.90 \\
          \rowcolor{Gray}
          T5\textsubscript{Large} & 91.93\textsubscript{0.32} & 91.39\textsubscript{0.75} & -0.59 \\
          RoBERTa\textsubscript{Large} & 92.71\textsubscript{0.21} & 93.30\textsubscript{0.24} & +0.64 \\
          \rowcolor{Gray}
          Longformer\textsubscript{Base} & 91.78\textsubscript{0.47} & 92.70\textsubscript{0.16} & +1.00 \\
          news-RoBERTa\textsubscript{Base} & 91.81\textsubscript{0.55} & 92.78\textsubscript{0.40} & +1.06 \\
          \rowcolor{Gray}
          Luke & \textbf{94.25}\textsubscript{0.21} & \textbf{95.29}\textsubscript{0.37} & \textbf{+1.10} \\
          \hline
      \end{tabular}
  }
  \caption{Detailed performances of the models on the CoNLL-2003 test set and the CoNLL++ test set, ranked
  by the $\Delta$F\textsubscript{1}.
  The performances are F\textsubscript{1} scores calculated by taking the average over five runs and the standard
  deviations are presented in subscripts. The best results are highlighted in bold.}
  \label{tab:detail_results}
\end{table}

\subsection{Model Size}
\label{ssec:app_model_size}
Table~\ref{tab:model_size} includes the results from Section~\ref{ssec:effect_of_size},
showing the performance statistics of BERT-based, ALBERT-based, RoBERTa-based, and T5-based
models with various sizes on the CoNLL-2003 and CoNLL++ test set. One side note is that
our results also confirms the previous findings that the performance on a downstream task
has a positive correlation with the model size.

\begin{table}[h!]
  \resizebox{\columnwidth}{!}{
    % \centering
    % \tiny
    \begin{tabular}{|l|c|c|c|c|}
      \hline
      \textbf{Name} & \textbf{\# Parameters} & \textbf{CoNLL-2003} & \textbf{CoNLL++} & \textbf{$\Delta$F\textsubscript{1} (\%)} \\
      \hline
      \rowcolor{Gray}
      BERT\textsubscript{Base} & 108M & 91.38\textsubscript{0.33} & 87.73\textsubscript{0.51} & -3.99 \\
      \rowcolor{Gray}
      BERT\textsubscript{Large} & 334M & 91.77\textsubscript{0.20} & 89.93\textsubscript{0.74} & -2.01 \\
      RoBERTa\textsubscript{Base} & 123M & 92.08\textsubscript{0.22} & 93.13\textsubscript{0.31} & +1.14 \\
      RoBERTa\textsubscript{Large} & 354M & 92.71\textsubscript{0.21} & 93.30\textsubscript{0.24} & +0.64 \\
      \rowcolor{Gray}
      ALBERT\textsubscript{Base} & 12M & 89.53\textsubscript{0.23} & 86.30\textsubscript{0.39} & -3.61 \\
      \rowcolor{Gray}
      ALBERT\textsubscript{Large} & 18M & 90.46\textsubscript{0.21} & 87.44\textsubscript{0.47} & -3.34 \\
      \rowcolor{Gray}
      ALBERT\textsubscript{XLarge} & 60M & 90.80\textsubscript{0.17} & 88.57\textsubscript{1.03} & -2.46 \\
      \rowcolor{Gray}
      ALBERT\textsubscript{XXLarge} & 235M & 91.69\textsubscript{0.33} & 89.65\textsubscript{0.23} & -2.22 \\
      T5\textsubscript{Small} & 60M & 88.94\textsubscript{0.32} & 86.36\textsubscript{0.08} & -2.90 \\
      T5\textsubscript{Base} & 220M & 91.55\textsubscript{0.27} & 90.05\textsubscript{0.45} & -1.64 \\
      T5\textsubscript{Large} & 770M & 91.93\textsubscript{0.32} & 91.39\textsubscript{0.75} & -0.59 \\
      T5\textsubscript{3B} & 3B & 92.59\textsubscript{0.32} & 93.21\textsubscript{0.09} & +0.67 \\
      \hline
      % T5\textsubscript{11B} & x & x & x
    \end{tabular}
  }
  \caption{Performances of the models of different sizes on the CoNLL-2003 test set and the CoNLL++ test set.
  The performances are F\textsubscript{1} scores calculated by taking the average over five runs and the standard
  deviations are presented in subscripts.}
  \label{tab:model_size}
\end{table}

\subsection{Number of Fine-Tuning Examples}
\label{ssec:app_num_examples}
Table~\ref{tab:example_no_roberta} and Table~\ref{tab:example_no_flair} show the results
from Section~\ref{ssec:amount_of_training_examples} of the RoBERTa-based and Flair-based
models on the two test sets respectively when varying the number of examples fine-tuned on.

\begin{table}[h!]
  \resizebox{\columnwidth}{!}{
    % \centering
    % \tiny
    \begin{tabular}{|l|c|c|c|}
      \hline
      \begin{tabular}{@{}c@{}} \textbf{Training} \\ \textbf{Example} \end{tabular} & \textbf{CoNLL-2003} & \textbf{CoNLL++} & \textbf{$\Delta$F\textsubscript{1} (\%)} \\
      \hline
      10\% & 88.28\textsubscript{0.38} & 88.49\textsubscript{0.67} & +0.24 \\
      \rowcolor{Gray}
      20\% & 90.23\textsubscript{0.30} & 91.08\textsubscript{0.47} & +0.94 \\
      30\% & 90.81\textsubscript{0.21} & 91.36\textsubscript{0.40} & +0.61 \\
      \rowcolor{Gray}
      40\% & 91.10\textsubscript{0.12} & 91.64\textsubscript{0.48} & +0.59 \\
      50\% & 91.42\textsubscript{0.15} & 91.76\textsubscript{0.49} & +0.37 \\
      \rowcolor{Gray}
      60\% & 91.45\textsubscript{0.27} & 91.93\textsubscript{0.34} & +0.52 \\
      70\% & 91.82\textsubscript{0.10} & 92.25\textsubscript{0.34} & +0.47 \\
      \rowcolor{Gray}
      80\% & 91.98\textsubscript{0.15} & 92.97\textsubscript{0.46} & +1.07 \\
      90\% & 92.04\textsubscript{0.20} & 92.94\textsubscript{0.50} & +0.98 \\
      \hline
    \end{tabular}
    \caption{Performances of RoBERTa\textsubscript{Base} on the CoNLL-2003 test set and the CoNLL++ test set when varying the percentage of training examples used.
    The performances are F\textsubscript{1} scores calculated by taking the average over five runs and the standard
    deviations are presented in subscripts.}
    \label{tab:example_no_roberta}
  }
\end{table}

\begin{table}[h!]
  \resizebox{\columnwidth}{!}{
  \begin{tabular}{|l|c|c|c|}
    \hline
    \begin{tabular}{@{}c@{}} \textbf{Training} \\ \textbf{Example} \end{tabular} & \textbf{CoNLL-2003} & \textbf{CoNLL++} & \textbf{$\Delta$F\textsubscript{1} (\%)} \\
    \hline
    10\% & 86.90\textsubscript{0.15} & 79.11\textsubscript{0.53} & -8.96 \\
    \rowcolor{Gray}
    20\% & 88.42\textsubscript{0.45} & 82.26\textsubscript{0.67} & -6.96 \\
    30\% & 89.04\textsubscript{0.24} & 83.17\textsubscript{0.71} & -6.59 \\
    \rowcolor{Gray}
    40\% & 89.74\textsubscript{0.11} & 83.98\textsubscript{0.49} & -6.43 \\
    50\% & 90.15\textsubscript{0.13} & 84.47\textsubscript{0.26} & -6.30 \\
    \rowcolor{Gray}
    60\% & 90.40\textsubscript{0.28} & 84.64\textsubscript{0.74} & -6.38 \\
    70\% & 90.62\textsubscript{0.16} & 85.08\textsubscript{0.83} & -6.11 \\
    \rowcolor{Gray}
    80\% & 90.68\textsubscript{0.16} & 85.39\textsubscript{0.62} & -5.83 \\
    90\% & 90.84\textsubscript{0.17} & 85.44\textsubscript{0.46} & -5.94 \\
    \hline
  \end{tabular}
  \caption{Performances of Flair on the CoNLL-2003 test set and the CoNLL++ test set when varying the percentage of training examples used.}
  \label{tab:example_no_flair}
  }
\end{table}

\subsection{Temporal Drift}
\label{ssec:app_temporal_drift}
Table~\ref{tab:temporal_drift} show the results
from Section~\ref{ssec:temporal_drift}. The ``Year" column indicates the time period from which the data used for continued pre-training on RoBERTa\textsubscript{Base} was used.

\begin{table}[h!]
  \resizebox{\columnwidth}{!}{
  \begin{tabular}{|l|c|c|c|}
    \hline
    \textbf{Year} & \textbf{CoNLL-2003} & \textbf{CoNLL++} & \textbf{$\Delta$F\textsubscript{1} (\%)} \\
    \hline
    2007 & 91.96\textsubscript{0.44} & 92.85\textsubscript{0.31} & +0.97 \\
    \rowcolor{Gray}
    2008 & 91.88\textsubscript{0.09} & 92.69\textsubscript{0.17} & +0.88 \\
    2009 & 92.24\textsubscript{0.17} & 93.10\textsubscript{0.11} & +0.87 \\
    \rowcolor{Gray}
    2010 & 91.92\textsubscript{0.25} & 93.10\textsubscript{0.41} & +1.28 \\
    2011 & 92.07\textsubscript{0.35} & 92.95\textsubscript{0.15} & +0.96 \\
    \rowcolor{Gray}
    2012 & 92.07\textsubscript{0.34} & 92.77\textsubscript{0.33} & +0.76 \\
    2013 & 91.87\textsubscript{0.23} & 92.84\textsubscript{0.27} & +1.05 \\
    \rowcolor{Gray}
    2014 & 92.01\textsubscript{0.32} & 92.89\textsubscript{0.21} & +0.96 \\
    2015 & 91.95\textsubscript{0.29} & 92.92\textsubscript{0.63} & +0.99 \\
    \rowcolor{Gray}
    2016 & 91.98\textsubscript{0.23} & 92.92\textsubscript{0.26} & +1.02 \\
    2017 & 91.93\textsubscript{0.13} & 92.93\textsubscript{0.18} & +1.08 \\
    \rowcolor{Gray}
    2018 & 91.89\textsubscript{0.38} & 92.93\textsubscript{0.44} & +1.13 \\
    2019 & 91.80\textsubscript{0.29} & 93.25\textsubscript{0.44} & +1.58 \\
    \hline
  \end{tabular}
  \caption{Performances of differnt checkpoints obtained by continued pre-training RoBERTa\textsubscript{Base} with data from different years on the CoNLL-2003 test set and CoNLL++.}
  \label{tab:temporal_drift}
  }
\end{table}

\subsection{Test Reuse}
\label{ssec:app_test_reuse}
Table~\ref{tab:test_reuse_stats} shows the results from Section~\ref{sssec:test_reuse}.

\begin{table}[ht!]
  \resizebox{\columnwidth}{!}{
      \centering
      \tiny
      \begin{tabular}{|l|c|c|c|}
          \hline
          \textbf{Name} & \textbf{CoNLL-2003} & \textbf{CoNLL-2003'} & \textbf{$\Delta$F\textsubscript{1} (\%)} \\
          \hline
          \rowcolor{Gray}
          BiLSTM-CRF & 91.00\textsubscript{0.18} & 93.30\textsubscript{0.19} & +2.53 \\
          BiLSTM-CNN & 89.02\textsubscript{0.09} & 90.58\textsubscript{0.57} & +1.75 \\
          \rowcolor{Gray}
          % Stanford NLP & 88.28 & x & x \\
          SciBERT & 87.05\textsubscript{0.91} & 86.97\textsubscript{0.84} & -0.09 \\
          BiLSTM-CNN-CRF & 90.25\textsubscript{0.22} & 92.61\textsubscript{0.29} & +2.95 \\
          \rowcolor{Gray}
          % SpanBERT\textsubscript{Large} & 88.31\textsubscript{0.41} & 88.49\textsubscript{0.86} & +0.20 \\
          BiLSTM-CRF-ELMo & 92.36\textsubscript{0.10} & 93.82\textsubscript{0.07} & +1.58 \\
          Flair & 92.46\textsubscript{0.14} & 93.16\textsubscript{0.13} & +0.76 \\
          \rowcolor{Gray}
          % \rowcolor{Gray}
          % Stanza & 91.50 & x & x \\
          Pooled Flair & 93.15\textsubscript{0.24} & 94.64\textsubscript{0.09} & +1.60 \\
          mBERT & 91.06\textsubscript{0.42} & 93.61\textsubscript{0.33} & +2.80 \\
          \rowcolor{Gray}
          GigaBERT & 91.35\textsubscript{0.27} & 92.95\textsubscript{0.84} & +1.75 \\
          ALBERT\textsubscript{Base} & 89.53\textsubscript{0.23} & 91.64\textsubscript{0.17} & +2.36 \\
          \rowcolor{Gray}
          BERT\textsubscript{Large} & 91.77\textsubscript{0.20} & 92.20\textsubscript{0.85} & +0.47 \\
          XLM-RoBERTa\textsubscript{Base} & 91.04\textsubscript{0.53} & 89.66\textsubscript{9.42} & -1.52 \\
          \rowcolor{Gray}
          T5\textsubscript{Large} & 91.93\textsubscript{0.32} & 94.37\textsubscript{0.32} & +2.65 \\
          RoBERTa\textsubscript{Large} & 92.71\textsubscript{0.21} & 93.06\textsubscript{0.63} & +0.38 \\
          \rowcolor{Gray}
          Longformer\textsubscript{Base} & 91.78\textsubscript{0.47} & 94.02\textsubscript{0.48} & +2.44 \\
          news-RoBERTa\textsubscript{Base} & 91.81\textsubscript{0.55} & 93.55\textsubscript{0.23} & +1.90 \\
          \rowcolor{Gray}
          Luke & \textbf{94.25}\textsubscript{0.21} & \textbf{96.01}\textsubscript{0.15} & +1.87 \\
          \hline
      \end{tabular}
  }
  \caption{Detailed F\textsubscript{1} scores of the models on the CoNLL-2003 and the CoNLL-2003' test set.
  }
  \label{tab:test_reuse_stats}
\end{table}

% This is a section in the appendix.

\end{document}